\pdfoutput=1

\documentclass[11pt]{article}

\usepackage[]{acl}

\usepackage{times}
\usepackage{latexsym}

\usepackage[T1]{fontenc}

\usepackage[utf8]{inputenc}
\usepackage{inconsolata}
\usepackage{graphicx} 
\usepackage{booktabs}

\usepackage{microtype}

\title{Who Are All The Stochastic Parrots Imitating? They Should Tell Us!}

\author{Sagi Shaier$^\nabla$ \quad~\quad Lawrence E. Hunter$^\dag$ \quad~\quad Katharina von der Wense$^{\nabla\diamondsuit\spadesuit}$ \\
  $^\nabla$University of Colorado Boulder\\
$^\dag$University of Colorado Denver\\
$^\diamondsuit$Johannes Gutenberg University Mainz\\
$^\nabla$\texttt{\{sagi.shaier, katharina.kann\}@colorado.edu} \\
$^\dag$\texttt{larry.hunter@cuanschutz.edu}}

\begin{document}
\maketitle
\def\thefootnote{$\spadesuit$}\footnotetext{Formerly: Katharina Kann}
\begin{abstract}
Both standalone language models (LMs) as well as LMs within downstream-task systems have been shown to generate statements which are factually untrue. This problem is especially severe for low-resource languages, where training data is scarce and of worse quality than for high-resource languages. In this opinion piece, we argue that LMs in their current state will never be fully trustworthy in critical settings and suggest a possible novel strategy to handle this issue: by building LMs such that can cite their sources -- i.e., point a user to the parts of their training data that back up their outputs. We first discuss which current NLP tasks would or would not benefit from such models. We then highlight the expected benefits such models would bring, e.g., quick verifiability of statements. We end by outlining the individual tasks that would need to be solved on the way to developing LMs with the ability to cite. We hope to start a discussion about the field's current approach to building LMs, especially for low-resource languages, and the role of the training data in explaining model generations.
\end{abstract}

\section{Introduction}

Transformers \cite{attention} and related models have been improving rapidly,
with applications in a surprisingly large number of domains, such as natural language generation \cite{zhang-etal-2019-pretraining}, machine translation \cite{wang-etal-2019-learning-deep}, question answering \cite{akermi-etal-2020-tansformer}, 
and code generation \cite{code_gen}, based on the ability to generate sensible outputs to prompts over a nearly limitless input domain. 

\begin{figure}[th]
    \includegraphics[width=1\columnwidth,height=0.7\columnwidth,keepaspectratio]{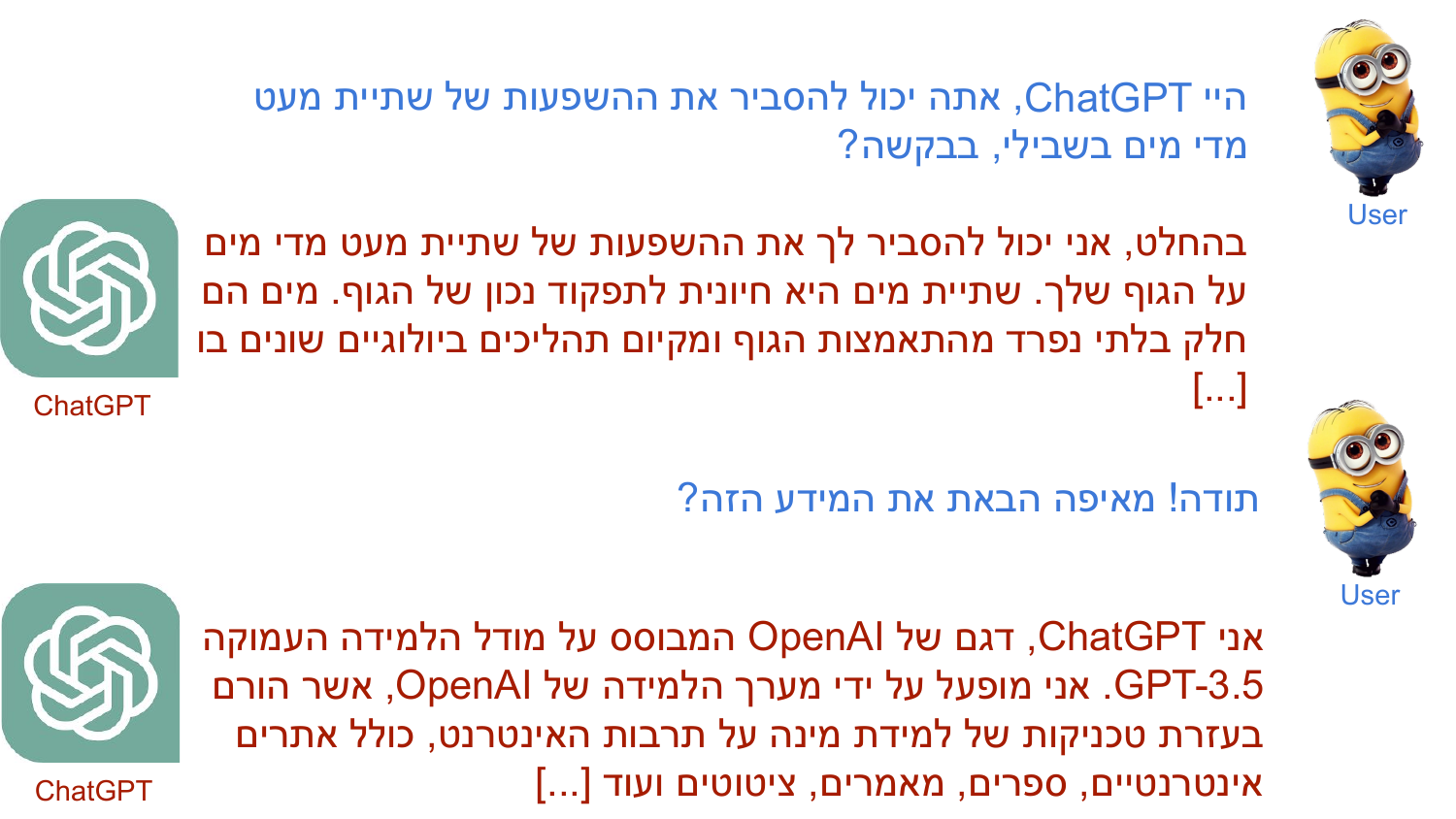}
    \caption{An actual conversation with ChatGPT in Hebrew on the effects of not drinking enough water. ChatGPT is unable to point the user to its sources and instead falls back to a general answer (“I am ChatGPT, an OpenAI model based on the GPT-3.5 deep learning model. I am powered by OpenAI's learning set, which has been raised with the help of machine learning techniques on Internet culture, including websites, books, articles, quotes, and more”). We argue that ChatGPT and similar models should be able to direct the user to the sources of their information, which will have multiple benefits, such as quick verifiability of model statements.}
    \label{main_figure}
\end{figure}

Despite impressive performance on a wide array of benchmark tasks, these models are known to produce “AI-splaining,” confident sounding but incorrect statements: “To the extent that a use case places importance on the truth of the outputs provided, it is not a good fit for GPT-3” \cite{dale_2021}; see also \citet{church-etal-2022-gentle} and 
\citet{Rebooting}. 

This problem has proven to be especially true for models trained on low-resource languages \cite{guerreiro2023hallucinations}, where data may not only be scarce \cite{mager-etal-2018-challenges}, but also not well curated with respect to correctness or quality, in comparison to higher-resource languages \cite{hedderich-etal-2021-survey}. Furthermore, model hallucination in such settings can result in toxic patterns that can be found in the training data \cite{guerreiro2023hallucinations}.

In accordance with the large LMs and low-resource languages theme track, we argue that while the performance and factuality of LMs has been improving, both in high-resource and low-resource settings, in their existing state, LMs will realistically never be fully trustworthy. Thus, in settings in which factuality is required, such as medicine, they are dangerous and unemployable. This is further noted in \citet{gopher}, who state that users cannot trust any claim a model makes without fact-checking.

Our proposal to address these concerns suggests both technical development and a simple regulatory framework: as we often ask students, journalists and scholars, \textbf{we should ask our models to name their sources and provide evidence for their assertions.} Currently, even popular LMs often fail at this, as seen on the ChatGPT \cite{openai_2023} example in Figure \ref{main_figure}.
In the case of generative models, either the model itself or a post-hoc procedure could -- and, under certain circumstances, \emph{should be required to} -- be designed to produce evidentiary justification for its output.  

NLP tasks would benefit from such citation models,
discuss the benefits 
they would bring, 
and present a roadmap to develop such models. 
Our goal is to motivate the field to start thinking about what is 
necessary to make current models truly useful in all sorts of -- potentially critical -- scenarios.

\section{Background}
\paragraph{Factuality and the Lack Thereof}
LMs store factual knowledge \cite{factualKnowledge, shaier-etal-2023-emerging, editing_fact, measuring_fact, know_lms} and previous work have shown that LMs can act as knowledge bases \cite{LMsKBs, biolama}. However, there is no guarantee that the retrieved knowledge 
is indeed factual, and unfortunately, often it is not. This can be seen in many areas, such as question answering \cite{qa_hallu}, dialogue systems \cite{dziri-etal-2021-neural, retrieval_hall, testoni-bernardi-2021-ive}, image captioning \cite{rohrbach-etal-2018-object}, 
text summarization \cite{zhao-etal-2020-reducing, cao-etal-2022-hallucinated, maynez-etal-2020-faithfulness} and translation \cite{cur_case, hall_sum}. This is especially true in low-resource settings \cite{guerreiro2023hallucinations}. In order for LMs to be fully utilized as such knowledge bases and in settings where factuality is crucial, the retrieved knowledge must first be factual. But, without knowing the source of such the model's knowledge, verifying its factuality is a challenge.

\paragraph{Citation Generation}
\label{Citation_Generation}
Although LMs, particularly those intended to produce scientific text, such as Meta's Galactica \cite{galactica}, already produce text that looks as if it is a citation, frequently there is no document corresponding to the apparent citation or the cited document does not support the statement associated with it. Many existing approaches to citation recommendation offer productive avenues to explore for factuality testing, post-hoc generation of support, hybrid architectures, or creation of training data \cite{ali_cit, text_sampling}. There has also been work on citation generation, where the task is either: 1) given two documents, generate an explanation for the relation between them \cite{luu2020citation}, or 2) generate a citation for an already existing text \cite{control_cit_gen, xing-etal-2020-automatic, cit_gen_control, Fetahu_2016}. This is different from our suggestion to generate statements and citations simultaneously, and also not optimal: as LMs are being trained on massive datasets, evaluating whether each statement came from each of the potentially millions of article becomes impractical. Lastly, many existing systems that can in fact provide citations are based on search engines or retrieval models \cite{gopher, sparrow}, see also Perplexity AI\footnote{https://www.perplexity.ai/}, YouChat\footnote{https://you.com/}, or the ALCE benchmark \cite{gao2023enabling}. This is problematic because 1) it is far more time consuming than directly generating citations together with text; 2) access to the information sources needs to be provided at all times; 3) in contrast to our proposed approach, it does not increase model interpretability; and 4) for low-resource languages the quantity and quality of the data is often limited, and hence result in difficulties retrieving the relevant, factual source.

\section{Citations and Their Pros and Cons}
\label{cit_requirement}

In this section, we will first discuss which NLP tasks -- according to us -- require LMs with an ability to cite their sources. We will then discuss the benefits and, subsequently, risks of such models.

\subsection{Which Tasks Require Citations?}
\label{which_task}
We propose to classify tasks via two questions: (1) Is the source of the generated text obvious? (2) Is the generated text an objective truth or a subjective statement? See Table \ref{citation-task-table} for examples. 

If the answer to the first question is \textit{yes}, no further citation is required. This is the case, e.g., for machine translation \cite{brants-etal-2007-large}: the content of the generated text comes from the input sentence. The same holds true for summarization \cite{see-etal-2017-get} and paraphrase generation \cite{zhou-bhat-2021-paraphrase}. However, this is only partially the case for text simplification \cite{sheang-saggion-2021-controllable}: while most of the content comes from the original text, simpler versions of text sometimes contain additional explanations, which do require citations. 
In contrast, for many other tasks the input does not act as the source for text generation -- instead, the output comes from information stored in the model parameters and, thus, originally from the training data. An ideal system would be able to cite the part of its training data responsible for any given output. This is the case for the popular NLP tasks of closed-book free-text question answering \cite{roberts-etal-2020-much}, dialogue generation \cite{zhao-etal-2020-knowledge-grounded}, or creative writing \cite{story}. 

For tasks for which the answer to Question 1 is \textit{no}, we then turn to the second aforementioned question and ask if the generated text without clear sources of information in the input contains what should be objective truths. This is typically true for closed-book free-text question answering, which, as a consequence, according to our rules does require citations. However, this is \textit{sometimes} the case for other tasks too, such as the generation of additional explanations during text simplification or image captioning.
Similarly, for dialogue generation, objective truths and subjective statements could be mixed within the same conversation. As a result, some generated statements for those tasks do require citations, while others are good without.

\begin{table*}[t]
\centering\small\setlength{\tabcolsep}{2.5pt}
\centering
\small
\begin{tabular}{ccccp{3in}}
\toprule
\textbf{Task} &  \textbf{Q1} & \textbf{Q2} & \textbf{Citation?} & \textbf{Example} \\ \midrule
Creative writing                & No & Sometimes & Sometimes               &  Penguins are known for their ability to survive in harsh Antarctic conditions [CITATION], but few people know that they also possess the power of telekinesis which they use to build intricate nests out of ice blocks.
  \\ 
Dialogue generation                  & No        & Sometimes & Sometimes              &  

Did you know that penguins can jump up to 6 feet out of water when leaping onto land or ice floes? [CITATION]. I think elephants can do the same.
  \\ 
Free-text QA          & No   & Yes & Yes                    & The current president is not a penguin [CITATION]. \\ 
Image captioning & No  &Sometimes & Sometimes                &   
A group of penguins diving into the ocean to catch fresh fish for dinner, highlighting their impressive swimming abilities [CITATION], while one penguin emerges victorious with a giant fish twice its size.

   \\ 
Paraphrase generation          & Yes      & N/A & No             &   \textbf{Source text}: Penguins are social animals who live in large colonies.
\textbf{Paraphrased sentence}: Penguins thrive in community living  \\ 
Summarization           & Yes        & N/A & No            &  
\textbf{Source text}: Emperor penguins are the largest species of penguin, standing up to 4 feet tall. They are skilled hunters, capable of catching fish and krill by diving hundreds of feet below the surface. \textbf{Summary}: Emperor penguins are notable for their size and hunting prowess, making them formidable predators in their environment.

  \\ 
Text simplification             & Sometimes   & Sometimes & Sometimes               & \textbf{Source text}: Penguins have evolved unique adaptations that allow them to survive in environments as harsh as Antarctica, such as their countershaded dark and white plumage, which camouflages them from predators above and below the ice. \textbf{Simplified text}: Penguins live in Antarctica, which is year-round one of the coldest places on Earth [CITATION], and they look different than other birds so they don't get eaten.
   \\ 
Translation       & Yes         & N/A & No            &   \textbf{Source text}: Penguins are cool. \textbf{Translated text}: Pinguine sind cool.  \\ \bottomrule
\end{tabular}
\caption{An overview of natural language generation tasks together with our opinion regarding if they require citations. Q1: \textit{Obvious source?} Q2: \textit{Objective truth?}}
\label{citation-task-table}
\end{table*}

\subsection{Benefits of Citations}
\label{benefits}

Citations allow us to verify the factuality of generated text easily. In contrast, without knowing where the text came from we are often unable to verify that it is correct. Moreover, knowing what portion of the text is copied verbatim allows us to give credit to the author and prevent copyright violations. Citations also increase the explainability of the answer and allow users to learn more about interesting topics. 

Additionally, recent work in prompt engineering have shown that models providing justifications for their assertions (even when only partially correct) can improve the correctness of the outputs \cite{jung}. Trustworthiness judgments among people often include a social aspect, so by doing a good job of identifying sources and influences has the potential to increase both the trust in AI systems and their trustworthiness.   For example, human trustworthiness judgments about scientific claims are influenced by the interests of the authors \cite{Beware}.

\subsection{Risks of Citations}
\label{risks}
Unfortunately, citations also come with risks. Just by having a citation next to a generated text, users are more likely to trust it \cite{Thornley2015TheRO}. However, it is likely that users will not examine each and every citation manually to verify that the text is indeed factual, or that the source is trustworthy \cite{read, Thornley2015TheRO}. This will be exacerbated by the fact that it is incredibly unlikely that any automated system will ever produce $100\%$ correct citations at all times, and may result in either users' diminishing trust and usage of such systems or a potential harm.

There is also the risk of decreased readability: backing up every statement with many citations, as the text may appear in multiple places, will reduce the readability of the text and may hinder users from reading or understanding it. Lastly, privacy concerns also arise from the training process of LMs. For example, state of the art LMs are often trained on a massive automatically extracted text \cite{lm_data}. But, as manual examination of each text is not feasible for its size, there is a possibility that it may contain private user information, such as patient records. This may result in LMs cite information that should stay private.

\subsection{Citations vs. Explainability}
The goal to understand why a model generates any given output is shared with research on model explainability \cite{danilevsky-etal-2020-survey}. However, in contrast to the latter, we are not interested in the effect of certain input on the output. In addition, we do not necessarily require that the model describes its reasoning by providing citations -- what we care about instead is that the citations back up the model's answer. This enables humans to verify the output -- even if the cited source should not actually in the technical sense have been the reason for the model's output.

\section{Road Map}
\label{roadmap}

\subsection{The Big Picture}
\label{BigPicture}
\paragraph{Meta-information} Currently, the standard in the field is to train models on text, disjoint from its origin. Even though some models are trained on data that contain text with citations (e.g., \citet{galactica}), the citations are only "attached" to statements taken from other sources, while any other text, even taken from the same article, does not have a citation attached to it. This results in LMs that can only sometimes, on a limited text, produce citations. In order to develop LMs that can cite their sources effectively, we need to give them the metadata which contain citation information. 

\paragraph{Retrieval} Say we trained a LM with the right data such that it has knowledge of which statement came from which article. How would we extract text with citations? One avenue for such knowledge extraction is to modify the pretraining, such that citation information is being generated together with every piece of generated text. 

\paragraph{When To Cite?} The above strategy would result in LMs that would always produce a citation. However, as mentioned in Section \ref{which_task}, not every task or statement requires a citation. For tasks that do require citations, we can just let the model always cite. For tasks that do not require citations, we can simply remove the citations. For tasks in between, where citation is sometimes required, we propose to utilize the existing subjectivity classification task \cite{wiebe-etal-1999-development}.

\subsection{Concrete Tasks to Master}
Our goal is to lay out a roadmap for the community, which describes necessary steps for the development of models that can cite their sources. This is not trivial, as it requires improvements of models for existing tasks as well as the development of systems for novel challenges.

\paragraph{Simultaneous Citation and Text Generation}
As mentioned in Section \ref{Citation_Generation}, existing work mainly retrieve citations for already generated text, which becomes intractable as models are trained on ever more text and the number of possible source documents increases drastically. In contrast, we propose STANCE: the task of \textbf{S}imultaneous \textbf{T}ext \textbf{AN}d \textbf{C}itation g\textbf{E}neration. As an additional challenge, future work should also focus on MultiSTANCE: multihop citation generation, where the sources for a given text are spread across multiple texts. As the number of citations can be significant (though much smaller in the low-resource setting), we suggest to use topic modeling, as a potential avenue to reduce such large search space.

\paragraph{Subjectivity Classification}
As mentioned in Section \ref{which_task}, whether a task requires a citation partially depends on if the text is objective or subjective. This is not a novel task as the community has been working on subjectivity classification for quite some time 
\cite{wiebe-etal-1999-development, Wiebe2005CreatingSA}. 
However, to the best of our knowledge, models for this task have not been employed in the context of citations. 

\paragraph{Citation-Text Correctness}
To ensure that the retrieval step (Section \ref{BigPicture}) is successful, we need to identify whether the statement appears in the source. For that, two existing tasks can be used: 1) identifying which part of the generated text refers to the citation \cite{8931592}. 2) Validate that the citation is appropriate for the selected text span \cite{fully_aut_, fact_check, true, fact_community, lm_fact}. Using such automatic methods instead of manually verifying citations will result in faster model development. 

\paragraph{Source Trustworthiness}
We all know that Wikipedia is not a reliable source for citation. We propose CUE (\textbf{C}itation q\textbf{U}ality \textbf{E}valuation), the task of evaluating the quality of the source corresponding to a generated citation. We believe there are six main sub-tasks for CUE, which consist of classifying 1) the time of publication, 2) whether the source is credible, 3) how many times the source has been cited, 4) if the author is known, 5) if the source is unbiased, and 6) if the statement and citation are still relevant. For example, answering that the current US president is Barack Obama was \emph{previously} factual, and may still show up in many source documents, but is not factual in 2023.

\section{Conclusion}
Language models (LMs) performance has been improving rapidly in a wide variety of areas. However, there is the crucial issue of their generated text often being nonfactual, especially for low-resource languages. We argue that, in order for LMs to be fully trustworthy, they must cite their sources -- i.e., point users to the parts of their training data that back up their outputs. In this opinion piece, we discuss NLP tasks which would benefit from such citation models, highlight the benefits and risks such models would bring, and outline the individual tasks that would need to be solved on the way to develop such LMs. 

\section*{Limitations}
While developing the proposed language models that can cite their sources increase their utility, there is a risk that people would trust them more without actually verifying that the generated citations are actually correct. Such increase in trust would be especially problematic in time-critical scenarios where people cannot examine each citation manually. Ideally, our proposal will result in an increase of data cleaning, such that each citation is by default trustworthy. That being said, our approach does not solve copyright issues.

\section*{Ethics Statement}
The main reason for this paper is to point out shortcomings of state-of-the-art language models, which can have significant social, health-related, and economic consequences. Future work should develop systems that can cite their sources in order to facilitate a verification of the factuality of generated statements.

\section*{Acknowledgments}
We would like to thank the reviewers for taking the time to provide feedback on our work. Your insights have been valuable in helping us refine and enhance our manuscript. The authors acknowledge financial support from NIH grants OT2TR003422 and R01LM013400.


\bibliography{anthology,custom}

\begin{thebibliography}{61}
\expandafter\ifx\csname natexlab\endcsname\relax\def\natexlab#1{#1}\fi

\bibitem[{Akermi et~al.(2020)Akermi, Heinecke, and Herledan}]{akermi-etal-2020-tansformer}
Imen Akermi, Johannes Heinecke, and Fr{\'e}d{\'e}ric Herledan. 2020.
\newblock \href {https://aclanthology.org/2020.inlg-1.41} {Transformer based natural language generation for question-answering}.
\newblock In \emph{Proceedings of the 13th International Conference on Natural Language Generation}, pages 349--359, Dublin, Ireland. Association for Computational Linguistics.

\bibitem[{Ali et~al.(2022)Ali, Qi, Muhammad, Bhattacharyya, Ullah, and Abro}]{ali_cit}
Zafar Ali, Guilin Qi, Khan Muhammad, Siddhartha Bhattacharyya, Irfan Ullah, and Waheed Abro. 2022.
\newblock \href {https://doi.org/10.1007/s00521-021-06135-y} {Citation recommendation employing heterogeneous bibliographic network embedding}.
\newblock \emph{Neural Comput. Appl.}, 34(13):10229–10242.

\bibitem[{Brants et~al.(2007)Brants, Popat, Xu, Och, and Dean}]{brants-etal-2007-large}
Thorsten Brants, Ashok~C. Popat, Peng Xu, Franz~J. Och, and Jeffrey Dean. 2007.
\newblock \href {https://aclanthology.org/D07-1090} {Large language models in machine translation}.
\newblock In \emph{Proceedings of the 2007 Joint Conference on Empirical Methods in Natural Language Processing and Computational Natural Language Learning ({EMNLP}-{C}o{NLL})}, pages 858--867, Prague, Czech Republic. Association for Computational Linguistics.

\bibitem[{Cao et~al.(2022)Cao, Dong, and Cheung}]{cao-etal-2022-hallucinated}
Meng Cao, Yue Dong, and Jackie Cheung. 2022.
\newblock \href {https://doi.org/10.18653/v1/2022.acl-long.236} {Hallucinated but factual! inspecting the factuality of hallucinations in abstractive summarization}.
\newblock In \emph{Proceedings of the 60th Annual Meeting of the Association for Computational Linguistics (Volume 1: Long Papers)}, pages 3340--3354, Dublin, Ireland. Association for Computational Linguistics.

\bibitem[{Church et~al.(2022)Church, Kordoni, Marcus, Davis, Ma, and Chen}]{church-etal-2022-gentle}
Kenneth Church, Valia Kordoni, Gary Marcus, Ernest Davis, Yanjun Ma, and Zeyu Chen. 2022.
\newblock \href {https://doi.org/10.18653/v1/2022.acl-tutorials.1} {A gentle introduction to deep nets and opportunities for the future}.
\newblock In \emph{Proceedings of the 60th Annual Meeting of the Association for Computational Linguistics: Tutorial Abstracts}, pages 1--6, Dublin, Ireland. Association for Computational Linguistics.

\bibitem[{Dale(2021)}]{dale_2021}
Robert Dale. 2021.
\newblock \href {https://doi.org/10.1017/S1351324920000601} {Gpt-3: What’s it good for?}
\newblock \emph{Natural Language Engineering}, 27(1):113–118.

\bibitem[{Danilevsky et~al.(2020)Danilevsky, Qian, Aharonov, Katsis, Kawas, and Sen}]{danilevsky-etal-2020-survey}
Marina Danilevsky, Kun Qian, Ranit Aharonov, Yannis Katsis, Ban Kawas, and Prithviraj Sen. 2020.
\newblock \href {https://aclanthology.org/2020.aacl-main.46} {A survey of the state of explainable {AI} for natural language processing}.
\newblock In \emph{Proceedings of the 1st Conference of the Asia-Pacific Chapter of the Association for Computational Linguistics and the 10th International Joint Conference on Natural Language Processing}, pages 447--459, Suzhou, China. Association for Computational Linguistics.

\bibitem[{De~Cao et~al.(2021)De~Cao, Aziz, and Titov}]{editing_fact}
Nicola De~Cao, Wilker Aziz, and Ivan Titov. 2021.
\newblock \href {https://doi.org/10.48550/ARXIV.2104.08164} {Editing factual knowledge in language models}.

\bibitem[{Dong et~al.(2022)Dong, Dai, Song, Xu, Sui, and Li}]{factualKnowledge}
Qingxiu Dong, Damai Dai, Yifan Song, Jingjing Xu, Zhifang Sui, and Lei Li. 2022.
\newblock \href {https://doi.org/10.48550/ARXIV.2210.03329} {Calibrating factual knowledge in pretrained language models}.

\bibitem[{Dziri et~al.(2021)Dziri, Madotto, Za{\"\i}ane, and Bose}]{dziri-etal-2021-neural}
Nouha Dziri, Andrea Madotto, Osmar Za{\"\i}ane, and Avishek~Joey Bose. 2021.
\newblock \href {https://doi.org/10.18653/v1/2021.emnlp-main.168} {Neural path hunter: Reducing hallucination in dialogue systems via path grounding}.
\newblock In \emph{Proceedings of the 2021 Conference on Empirical Methods in Natural Language Processing}, pages 2197--2214, Online and Punta Cana, Dominican Republic. Association for Computational Linguistics.

\bibitem[{Elazar et~al.(2021)Elazar, Kassner, Ravfogel, Ravichander, Hovy, Schütze, and Goldberg}]{measuring_fact}
Yanai Elazar, Nora Kassner, Shauli Ravfogel, Abhilasha Ravichander, Eduard Hovy, Hinrich Schütze, and Yoav Goldberg. 2021.
\newblock \href {https://doi.org/10.48550/ARXIV.2102.01017} {Measuring and improving consistency in pretrained language models}.

\bibitem[{Fetahu et~al.(2016)Fetahu, Markert, Nejdl, and Anand}]{Fetahu_2016}
Besnik Fetahu, Katja Markert, Wolfgang Nejdl, and Avishek Anand. 2016.
\newblock \href {https://doi.org/10.1145/2983323.2983808} {Finding news citations for wikipedia}.
\newblock In \emph{Proceedings of the 25th {ACM} International on Conference on Information and Knowledge Management}. {ACM}.

\bibitem[{Gao et~al.(2023)Gao, Yen, Yu, and Chen}]{gao2023enabling}
Tianyu Gao, Howard Yen, Jiatong Yu, and Danqi Chen. 2023.
\newblock \href {http://arxiv.org/abs/2305.14627} {Enabling large language models to generate text with citations}.

\bibitem[{Gierth and Bromme(2020)}]{Beware}
L.~Gierth and R.~Bromme. 2020.
\newblock {{B}eware of vested interests: {E}pistemic vigilance improves reasoning about scientific evidence (for some people)}.
\newblock \emph{PLoS One}, 15(4):e0231387.

\bibitem[{Glaese et~al.(2022)Glaese, McAleese, Trębacz, Aslanides, Firoiu, Ewalds, Rauh, Weidinger, Chadwick, Thacker, Campbell-Gillingham, Uesato, Huang, Comanescu, Yang, See, Dathathri, Greig, Chen, Fritz, Elias, Green, Mokrá, Fernando, Wu, Foley, Young, Gabriel, Isaac, Mellor, Hassabis, Kavukcuoglu, Hendricks, and Irving}]{sparrow}
Amelia Glaese, Nat McAleese, Maja Trębacz, John Aslanides, Vlad Firoiu, Timo Ewalds, Maribeth Rauh, Laura Weidinger, Martin Chadwick, Phoebe Thacker, Lucy Campbell-Gillingham, Jonathan Uesato, Po-Sen Huang, Ramona Comanescu, Fan Yang, Abigail See, Sumanth Dathathri, Rory Greig, Charlie Chen, Doug Fritz, Jaume~Sanchez Elias, Richard Green, Soňa Mokrá, Nicholas Fernando, Boxi Wu, Rachel Foley, Susannah Young, Iason Gabriel, William Isaac, John Mellor, Demis Hassabis, Koray Kavukcuoglu, Lisa~Anne Hendricks, and Geoffrey Irving. 2022.
\newblock \href {http://arxiv.org/abs/2209.14375} {Improving alignment of dialogue agents via targeted human judgements}.

\bibitem[{Gu and Hahnloser(2022)}]{control_cit_gen}
Nianlong Gu and Richard H.~R. Hahnloser. 2022.
\newblock \href {https://doi.org/10.48550/ARXIV.2211.07066} {Controllable citation text generation}.

\bibitem[{Guerreiro et~al.(2023)Guerreiro, Alves, Waldendorf, Haddow, Birch, Colombo, and Martins}]{guerreiro2023hallucinations}
Nuno~M. Guerreiro, Duarte Alves, Jonas Waldendorf, Barry Haddow, Alexandra Birch, Pierre Colombo, and André F.~T. Martins. 2023.
\newblock \href {http://arxiv.org/abs/2303.16104} {Hallucinations in large multilingual translation models}.

\bibitem[{Hedderich et~al.(2021)Hedderich, Lange, Adel, Str{\"o}tgen, and Klakow}]{hedderich-etal-2021-survey}
Michael~A. Hedderich, Lukas Lange, Heike Adel, Jannik Str{\"o}tgen, and Dietrich Klakow. 2021.
\newblock \href {https://doi.org/10.18653/v1/2021.naacl-main.201} {A survey on recent approaches for natural language processing in low-resource scenarios}.
\newblock In \emph{Proceedings of the 2021 Conference of the North American Chapter of the Association for Computational Linguistics: Human Language Technologies}, pages 2545--2568, Online. Association for Computational Linguistics.

\bibitem[{Honovich et~al.(2022)Honovich, Aharoni, Herzig, Taitelbaum, Kukliansy, Cohen, Scialom, Szpektor, Hassidim, and Matias}]{true}
Or~Honovich, Roee Aharoni, Jonathan Herzig, Hagai Taitelbaum, Doron Kukliansy, Vered Cohen, Thomas Scialom, Idan Szpektor, Avinatan Hassidim, and Yossi Matias. 2022.
\newblock \href {https://doi.org/10.48550/ARXIV.2204.04991} {True: Re-evaluating factual consistency evaluation}.

\bibitem[{Jeblick et~al.(2022)Jeblick, Schachtner, Dexl, Mittermeier, Stüber, Topalis, Weber, Wesp, Sabel, Ricke, and Ingrisch}]{hall_sum}
Katharina Jeblick, Balthasar Schachtner, Jakob Dexl, Andreas Mittermeier, Anna~Theresa Stüber, Johanna Topalis, Tobias Weber, Philipp Wesp, Bastian Sabel, Jens Ricke, and Michael Ingrisch. 2022.
\newblock \href {https://doi.org/10.48550/ARXIV.2212.14882} {Chatgpt makes medicine easy to swallow: An exploratory case study on simplified radiology reports}.

\bibitem[{Jiang et~al.(2019)Jiang, Xu, Araki, and Neubig}]{know_lms}
Zhengbao Jiang, Frank~F. Xu, Jun Araki, and Graham Neubig. 2019.
\newblock \href {https://doi.org/10.48550/ARXIV.1911.12543} {How can we know what language models know?}

\bibitem[{Jung et~al.(2022)Jung, Qin, Welleck, Brahman, Bhagavatula, Bras, and Choi}]{jung}
Jaehun Jung, Lianhui Qin, Sean Welleck, Faeze Brahman, Chandra Bhagavatula, Ronan~Le Bras, and Yejin Choi. 2022.
\newblock \href {https://doi.org/10.48550/ARXIV.2205.11822} {Maieutic prompting: Logically consistent reasoning with recursive explanations}.

\bibitem[{Karadzhov et~al.(2017)Karadzhov, Nakov, Marquez, Barron-Cedeno, and Koychev}]{fully_aut_}
Georgi Karadzhov, Preslav Nakov, Lluis Marquez, Alberto Barron-Cedeno, and Ivan Koychev. 2017.
\newblock \href {https://doi.org/10.48550/ARXIV.1710.00341} {Fully automated fact checking using external sources}.

\bibitem[{Krasnova et~al.(2023)Krasnova, Smaznevicha, and Baskakova}]{text_sampling}
F.~V. Krasnova, I.~S. Smaznevicha, and E.~N. Baskakova. 2023.
\newblock \href {https://doi.org/10.48550/ARXIV.2301.01673} {Text sampling strategies for predicting missing bibliographic links}.

\bibitem[{Lee et~al.(2020)Lee, Li, Wang, Yih, Ma, and Khabsa}]{lm_fact}
Nayeon Lee, Belinda~Z. Li, Sinong Wang, Wen-tau Yih, Hao Ma, and Madian Khabsa. 2020.
\newblock \href {https://doi.org/10.48550/ARXIV.2006.04102} {Language models as fact checkers?}

\bibitem[{Luu et~al.(2020)Luu, Koncel-Kedziorski, Lo, Cachola, and Smith}]{luu2020citation}
Kelvin Luu, Rik Koncel-Kedziorski, Kyle Lo, Isabel Cachola, and Noah~A Smith. 2020.
\newblock Citation text generation.
\newblock \emph{arXiv preprint arXiv:2002.00317}.

\bibitem[{Mager et~al.(2018)Mager, Gutierrez-Vasques, Sierra, and Meza-Ruiz}]{mager-etal-2018-challenges}
Manuel Mager, Ximena Gutierrez-Vasques, Gerardo Sierra, and Ivan Meza-Ruiz. 2018.
\newblock \href {https://aclanthology.org/C18-1006} {Challenges of language technologies for the indigenous languages of the {A}mericas}.
\newblock In \emph{Proceedings of the 27th International Conference on Computational Linguistics}, pages 55--69, Santa Fe, New Mexico, USA. Association for Computational Linguistics.

\bibitem[{Marcus(2019)}]{Rebooting}
Gary Ernest~Davis Marcus. 2019.
\newblock \emph{Rebooting AI: {B}uilding Artificial Intelligence we can Trust}.

\bibitem[{Martín et~al.(2021)Martín, Huertas-Tato, Huertas-García, Villar-Rodríguez, and Camacho}]{fact_check}
Alejandro Martín, Javier Huertas-Tato, Álvaro Huertas-García, Guillermo Villar-Rodríguez, and David Camacho. 2021.
\newblock \href {https://doi.org/10.48550/ARXIV.2110.14532} {Facter-check: Semi-automated fact-checking through semantic similarity and natural language inference}.

\bibitem[{Maynez et~al.(2020)Maynez, Narayan, Bohnet, and McDonald}]{maynez-etal-2020-faithfulness}
Joshua Maynez, Shashi Narayan, Bernd Bohnet, and Ryan McDonald. 2020.
\newblock \href {https://doi.org/10.18653/v1/2020.acl-main.173} {On faithfulness and factuality in abstractive summarization}.
\newblock In \emph{Proceedings of the 58th Annual Meeting of the Association for Computational Linguistics}, pages 1906--1919, Online. Association for Computational Linguistics.

\bibitem[{Menick et~al.(2022)Menick, Trebacz, Mikulik, Aslanides, Song, Chadwick, Glaese, Young, Campbell-Gillingham, Irving, and McAleese}]{gopher}
Jacob Menick, Maja Trebacz, Vladimir Mikulik, John Aslanides, Francis Song, Martin Chadwick, Mia Glaese, Susannah Young, Lucy Campbell-Gillingham, Geoffrey Irving, and Nat McAleese. 2022.
\newblock \href {http://arxiv.org/abs/2203.11147} {Teaching language models to support answers with verified quotes}.

\bibitem[{Mihaylova et~al.(2018)Mihaylova, Nakov, Marquez, Barron-Cedeno, Mohtarami, Karadzhov, and Glass}]{fact_community}
Tsvetomila Mihaylova, Preslav Nakov, Lluis Marquez, Alberto Barron-Cedeno, Mitra Mohtarami, Georgi Karadzhov, and James Glass. 2018.
\newblock \href {https://doi.org/10.48550/ARXIV.1803.03178} {Fact checking in community forums}.

\bibitem[{OpenAI(2023)}]{openai_2023}
OpenAI. 2023.
\newblock \href {https://openai.com/blog/chatgpt/} {Chatgpt: Optimizing language models for dialogue}.

\bibitem[{Petroni et~al.(2019)Petroni, Rocktäschel, Lewis, Bakhtin, Wu, Miller, and Riedel}]{LMsKBs}
Fabio Petroni, Tim Rocktäschel, Patrick Lewis, Anton Bakhtin, Yuxiang Wu, Alexander~H. Miller, and Sebastian Riedel. 2019.
\newblock \href {https://doi.org/10.48550/ARXIV.1909.01066} {Language models as knowledge bases?}

\bibitem[{Radford et~al.(2018)Radford, Wu, Child, Luan, Amodei, and Sutskever}]{lm_data}
Alec Radford, Jeffrey Wu, Rewon Child, David Luan, Dario Amodei, and Ilya Sutskever. 2018.
\newblock \href {https://d4mucfpksywv.cloudfront.net/better-language-models/language-models.pdf} {Language models are unsupervised multitask learners}.

\bibitem[{Raunak et~al.(2021)Raunak, Menezes, and Junczys-Dowmunt}]{cur_case}
Vikas Raunak, Arul Menezes, and Marcin Junczys-Dowmunt. 2021.
\newblock \href {https://doi.org/10.48550/ARXIV.2104.06683} {The curious case of hallucinations in neural machine translation}.

\bibitem[{Roberts et~al.(2020)Roberts, Raffel, and Shazeer}]{roberts-etal-2020-much}
Adam Roberts, Colin Raffel, and Noam Shazeer. 2020.
\newblock \href {https://doi.org/10.18653/v1/2020.emnlp-main.437} {How much knowledge can you pack into the parameters of a language model?}
\newblock In \emph{Proceedings of the 2020 Conference on Empirical Methods in Natural Language Processing (EMNLP)}, pages 5418--5426, Online. Association for Computational Linguistics.

\bibitem[{Rohrbach et~al.(2018)Rohrbach, Hendricks, Burns, Darrell, and Saenko}]{rohrbach-etal-2018-object}
Anna Rohrbach, Lisa~Anne Hendricks, Kaylee Burns, Trevor Darrell, and Kate Saenko. 2018.
\newblock \href {https://doi.org/10.18653/v1/D18-1437} {Object hallucination in image captioning}.
\newblock In \emph{Proceedings of the 2018 Conference on Empirical Methods in Natural Language Processing}, pages 4035--4045, Brussels, Belgium. Association for Computational Linguistics.

\bibitem[{See et~al.(2017)See, Liu, and Manning}]{see-etal-2017-get}
Abigail See, Peter~J. Liu, and Christopher~D. Manning. 2017.
\newblock \href {https://doi.org/10.18653/v1/P17-1099} {Get to the point: Summarization with pointer-generator networks}.
\newblock In \emph{Proceedings of the 55th Annual Meeting of the Association for Computational Linguistics (Volume 1: Long Papers)}, pages 1073--1083, Vancouver, Canada. Association for Computational Linguistics.

\bibitem[{Shaier et~al.(2023)Shaier, Bennett, Hunter, and Kann}]{shaier-etal-2023-emerging}
Sagi Shaier, Kevin Bennett, Lawrence Hunter, and Katharina Kann. 2023.
\newblock \href {https://aclanthology.org/2023.ijcnlp-main.36} {Emerging challenges in personalized medicine: Assessing demographic effects on biomedical question answering systems}.
\newblock In \emph{Proceedings of the 13th International Joint Conference on Natural Language Processing and the 3rd Conference of the Asia-Pacific Chapter of the Association for Computational Linguistics (Volume 1: Long Papers)}, pages 540--550, Nusa Dua, Bali. Association for Computational Linguistics.

\bibitem[{Sheang and Saggion(2021)}]{sheang-saggion-2021-controllable}
Kim~Cheng Sheang and Horacio Saggion. 2021.
\newblock \href {https://aclanthology.org/2021.inlg-1.38} {Controllable sentence simplification with a unified text-to-text transfer transformer}.
\newblock In \emph{Proceedings of the 14th International Conference on Natural Language Generation}, pages 341--352, Aberdeen, Scotland, UK. Association for Computational Linguistics.

\bibitem[{Shuster et~al.(2021)Shuster, Poff, Chen, Kiela, and Weston}]{retrieval_hall}
Kurt Shuster, Spencer Poff, Moya Chen, Douwe Kiela, and Jason Weston. 2021.
\newblock \href {https://doi.org/10.48550/ARXIV.2104.07567} {Retrieval augmentation reduces hallucination in conversation}.

\bibitem[{Simkin and Roychowdhury(2002)}]{read}
M.~V. Simkin and V.~P. Roychowdhury. 2002.
\newblock \href {https://doi.org/10.48550/ARXIV.COND-MAT/0212043} {Read before you cite!}

\bibitem[{Sung et~al.(2021)Sung, Lee, Yi, Jeon, Kim, and Kang}]{biolama}
Mujeen Sung, Jinhyuk Lee, Sean Yi, Minji Jeon, Sungdong Kim, and Jaewoo Kang. 2021.
\newblock \href {https://doi.org/10.48550/ARXIV.2109.07154} {Can language models be biomedical knowledge bases?}

\bibitem[{Svyatkovskiy et~al.(2020)Svyatkovskiy, Deng, Fu, and Sundaresan}]{code_gen}
Alexey Svyatkovskiy, Shao~Kun Deng, Shengyu Fu, and Neel Sundaresan. 2020.
\newblock \href {https://doi.org/10.48550/ARXIV.2005.08025} {Intellicode compose: Code generation using transformer}.

\bibitem[{Taylor et~al.(2022)Taylor, Kardas, Cucurull, Scialom, Hartshorn, Saravia, Poulton, Kerkez, and Stojnic}]{galactica}
Ross Taylor, Marcin Kardas, Guillem Cucurull, Thomas Scialom, Anthony Hartshorn, Elvis Saravia, Andrew Poulton, Viktor Kerkez, and Robert Stojnic. 2022.
\newblock \href {https://doi.org/10.48550/ARXIV.2211.09085} {Galactica: A large language model for science}.

\bibitem[{Testoni and Bernardi(2021)}]{testoni-bernardi-2021-ive}
Alberto Testoni and Raffaella Bernardi. 2021.
\newblock \href {https://doi.org/10.18653/v1/2021.acl-srw.11} {{``}{I}{'}ve seen things you people wouldn{'}t believe{''}: Hallucinating entities in {G}uess{W}hat?!}
\newblock In \emph{Proceedings of the 59th Annual Meeting of the Association for Computational Linguistics and the 11th International Joint Conference on Natural Language Processing: Student Research Workshop}, pages 101--111, Online. Association for Computational Linguistics.

\bibitem[{Thornley et~al.(2015)Thornley, Watkinson, Nicholas, Volentine, Jamali, Herman, Allard, Levine, and Tenopir}]{Thornley2015TheRO}
Clare Thornley, Anthony Watkinson, David Nicholas, Rachel Volentine, Hamid~R. Jamali, Eti Herman, Suzie Allard, Kenneth~J. Levine, and Carol Tenopir. 2015.
\newblock The role of trust and authority in the citation behaviour of researchers.
\newblock \emph{Inf. Res.}, 20.

\bibitem[{Vaswani et~al.(2017)Vaswani, Shazeer, Parmar, Uszkoreit, Jones, Gomez, Kaiser, and Polosukhin}]{attention}
Ashish Vaswani, Noam Shazeer, Niki Parmar, Jakob Uszkoreit, Llion Jones, Aidan~N. Gomez, Lukasz Kaiser, and Illia Polosukhin. 2017.
\newblock \href {https://doi.org/10.48550/ARXIV.1706.03762} {Attention is all you need}.

\bibitem[{Wang et~al.(2020)Wang, Li, Zhou, Tang, and Wang}]{8931592}
Pancheng Wang, Shasha Li, Haifang Zhou, Jintao Tang, and Ting Wang. 2020.
\newblock \href {https://doi.org/10.1109/ACCESS.2019.2959056} {Toc-rwg: Explore the combination of topic model and citation information for automatic related work generation}.
\newblock \emph{IEEE Access}, 8:13043--13055.

\bibitem[{Wang et~al.(2019)Wang, Li, Xiao, Zhu, Li, Wong, and Chao}]{wang-etal-2019-learning-deep}
Qiang Wang, Bei Li, Tong Xiao, Jingbo Zhu, Changliang Li, Derek~F. Wong, and Lidia~S. Chao. 2019.
\newblock \href {https://doi.org/10.18653/v1/P19-1176} {Learning deep transformer models for machine translation}.
\newblock In \emph{Proceedings of the 57th Annual Meeting of the Association for Computational Linguistics}, pages 1810--1822, Florence, Italy. Association for Computational Linguistics.

\bibitem[{Wiebe and Riloff(2005)}]{Wiebe2005CreatingSA}
Janyce Wiebe and Ellen Riloff. 2005.
\newblock Creating subjective and objective sentence classifiers from unannotated texts.
\newblock In \emph{Conference on Intelligent Text Processing and Computational Linguistics}.

\bibitem[{Wiebe et~al.(1999)Wiebe, Bruce, and O{'}Hara}]{wiebe-etal-1999-development}
Janyce~M. Wiebe, Rebecca~F. Bruce, and Thomas~P. O{'}Hara. 1999.
\newblock \href {https://doi.org/10.3115/1034678.1034721} {Development and use of a gold-standard data set for subjectivity classifications}.
\newblock In \emph{Proceedings of the 37th Annual Meeting of the Association for Computational Linguistics}, pages 246--253, College Park, Maryland, USA. Association for Computational Linguistics.

\bibitem[{Wu et~al.(2021)Wu, Shieh, Hsu, and Chen}]{cit_gen_control}
Jia-Yan Wu, Alexander Te-Wei Shieh, Shih-Ju Hsu, and Yun-Nung Chen. 2021.
\newblock \href {https://doi.org/10.48550/ARXIV.2112.01332} {Towards generating citation sentences for multiple references with intent control}.

\bibitem[{Xing et~al.(2020)Xing, Fan, and Wan}]{xing-etal-2020-automatic}
Xinyu Xing, Xiaosheng Fan, and Xiaojun Wan. 2020.
\newblock \href {https://doi.org/10.18653/v1/2020.acl-main.550} {Automatic generation of citation texts in scholarly papers: A pilot study}.
\newblock In \emph{Proceedings of the 58th Annual Meeting of the Association for Computational Linguistics}, pages 6181--6190, Online. Association for Computational Linguistics.

\bibitem[{Xu et~al.(2021)Xu, Liang, Huang, and Xiang}]{qa_hallu}
Peng Xu, Davis Liang, Zhiheng Huang, and Bing Xiang. 2021.
\newblock \href {https://doi.org/10.48550/ARXIV.2110.06393} {Attention-guided generative models for extractive question answering}.

\bibitem[{Xu et~al.(2020)Xu, Patwary, Shoeybi, Puri, Fung, Anandkumar, and Catanzaro}]{story}
Peng Xu, Mostofa Patwary, Mohammad Shoeybi, Raul Puri, Pascale Fung, Anima Anandkumar, and Bryan Catanzaro. 2020.
\newblock \href {https://doi.org/10.48550/ARXIV.2010.00840} {Megatron-cntrl: Controllable story generation with external knowledge using large-scale language models}.

\bibitem[{Zhang et~al.(2019)Zhang, Cai, Xu, and Wang}]{zhang-etal-2019-pretraining}
Haoyu Zhang, Jingjing Cai, Jianjun Xu, and Ji~Wang. 2019.
\newblock \href {https://doi.org/10.18653/v1/K19-1074} {Pretraining-based natural language generation for text summarization}.
\newblock In \emph{Proceedings of the 23rd Conference on Computational Natural Language Learning (CoNLL)}, pages 789--797, Hong Kong, China. Association for Computational Linguistics.

\bibitem[{Zhao et~al.(2020{\natexlab{a}})Zhao, Wu, Xu, Tao, Zhao, and Yan}]{zhao-etal-2020-knowledge-grounded}
Xueliang Zhao, Wei Wu, Can Xu, Chongyang Tao, Dongyan Zhao, and Rui Yan. 2020{\natexlab{a}}.
\newblock \href {https://doi.org/10.18653/v1/2020.emnlp-main.272} {Knowledge-grounded dialogue generation with pre-trained language models}.
\newblock In \emph{Proceedings of the 2020 Conference on Empirical Methods in Natural Language Processing (EMNLP)}, pages 3377--3390, Online. Association for Computational Linguistics.

\bibitem[{Zhao et~al.(2020{\natexlab{b}})Zhao, Cohen, and Webber}]{zhao-etal-2020-reducing}
Zheng Zhao, Shay~B. Cohen, and Bonnie Webber. 2020{\natexlab{b}}.
\newblock \href {https://doi.org/10.18653/v1/2020.findings-emnlp.203} {Reducing quantity hallucinations in abstractive summarization}.
\newblock In \emph{Findings of the Association for Computational Linguistics: EMNLP 2020}, pages 2237--2249, Online. Association for Computational Linguistics.

\bibitem[{Zhou and Bhat(2021)}]{zhou-bhat-2021-paraphrase}
Jianing Zhou and Suma Bhat. 2021.
\newblock \href {https://doi.org/10.18653/v1/2021.emnlp-main.414} {Paraphrase generation: A survey of the state of the art}.
\newblock In \emph{Proceedings of the 2021 Conference on Empirical Methods in Natural Language Processing}, pages 5075--5086, Online and Punta Cana, Dominican Republic. Association for Computational Linguistics.

\end{thebibliography}
\bibliographystyle{acl_natbib}
\end{document}